\documentclass[conference]{IEEEtran}
\IEEEoverridecommandlockouts
\usepackage{cite}
\usepackage{amsmath,amssymb,amsfonts}
\usepackage{algorithmic}
\usepackage{graphicx}
\usepackage{textcomp, multirow}
\usepackage{xcolor, comment}
\usepackage{cleveref, float}

\def\BibTeX{{\rm B\kern-.05em{\sc i\kern-.025em b}\kern-.08em
    T\kern-.1667em\lower.7ex\hbox{E}\kern-.125emX}}
\begin{document}

\title{Early Exiting Predictive Coding Neural Networks for Edge AI
 \\

\thanks{This research has received funding from the European Union’s Horizon research and innovation program under grant agreement No 101070374.} \thanks{\textcolor{red}{\textbf{This paper has been published in the 2025 33rd European Signal Processing Conference (doi: 10.23919/EUSIPCO63237.2025.11226813).}}}
}
 
\author{\IEEEauthorblockN{Alaa Zniber$^{{\star}}$, Mounir Ghogho$^{{\dagger}{\ddagger}}$, Ouassim Karrakchou$^{{\star}}$, Mehdi Zakroum$^{{\star}}$}
\IEEEauthorblockA{$^{\star}$ \textit{TICLab, International University of Rabat, Morocco}}
\IEEEauthorblockA{$^{\dagger}$ \textit{College of Computing, University Mohammed VI Polytechnic, Morocco}}
\IEEEauthorblockA{$^{\ddagger}$ \textit{School of Electronic and Electrical Engineering, University of Leeds, UK}}
\{alaa.zniber, ouassim.karrakchou, mehdi.zakroum\}@uir.ac.ma, mounir.ghogho@um6p.ma}

\maketitle

\begin{abstract}
The Internet of Things is transforming various fields, with sensors increasingly embedded in wearables, smart buildings, and connected equipment. While deep learning enables valuable insights from IoT data, conventional models are too computationally demanding for resource-limited edge devices. Moreover, privacy concerns and real-time processing needs make local computation a necessity over cloud-based solutions. 
Inspired by the brain’s energy efficiency, we propose a shallow bidirectional predictive coding network with early exiting, dynamically halting computations once a performance threshold is met. This reduces the memory footprint and computational overhead while maintaining high accuracy. We validate our approach using the CIFAR-10 dataset. Our model achieves performance comparable to deep networks with significantly fewer parameters and lower computational complexity, demonstrating the potential of biologically inspired architectures for efficient edge AI.
\end{abstract}


\section{Introduction}
\label{intro}

The Internet of Things (IoT) is transforming various domains, including health monitoring, smart cities, and precision agriculture, where vast amounts of data must be efficiently processed. To meet this demand, Deep Learning (DL) is integrated into IoT systems, enabling real-time monitoring, optimized actuation, and selective data storage \cite{iotdl}. However, privacy concerns and the need for low-latency processing drive a shift toward Edge AI, where inference is performed directly on IoT devices instead of cloud servers. This approach reduces dependence on remote infrastructure but introduces new challenges, as edge devices are often resource-constrained, making the deployment of large AI models difficult \cite{frugal}. On the one hand, DL models require substantial memory to store parameters and significant computational power to execute complex operations \cite{survey_extreme}. On the other hand, edge devices vary widely in computational resources, ranging from microcontrollers to cloudlets, including single-board computers \cite{murshed_machine_2021}. Hence, this variability poses a significant challenge, particularly for IoT extreme edge devices with only a few kilobytes of memory. In such cases, conventional DL models cannot be deployed directly without considerable modifications to their architecture.



Ensuring reliable inference on resource-constrained edge devices requires novel design approaches, and the human brain offers a compelling source of inspiration. Despite handling a vast array of tasks and processing large volumes of data, the brain operates with remarkable efficiency, consuming only around 20 Watts of power \cite{balasubramanian_brain_2021}. Indeed, the brain counts billions of neurons \cite{cell_nbr}, but its neuronal morphology could explain the observed frugality and efficacy \cite{van_bergen_going_2020}. The brain processes information through a structured hierarchy of layers, reinforced by bidirectional connections that enhance its ability to abstract and interpret the surrounding environment \cite{spoerer_recurrent_2020}. A key theoretical framework for understanding these interactions is predictive coding (PC). According to PC theory, the brain continuously minimizes prediction errors—the difference between actual and predicted stimuli—by refining its internal representations through feedback loops between higher and lower processing layers \cite{rao_predictive_1999}. 
When implemented with convolutional layers, where the feedforward path consists of convolution operators and the backward path of deconvolutions, 
Predictive Coding Networks (PCNs) often achieve higher accuracy than conventional architectures \cite{lotter_deep_2016, wen_deep_2018, local_pc} and exhibit greater robustness to adversarial attacks \cite{choksi_predify_2021}. 

However, existing PCNs have primarily focused on enhancing model performance rather than designing architectures optimized for low-resource devices.
Typically, PCNs are constructed by introducing feedback connections into deep feedforward networks and are trained for a {\em fixed} number of cycles. This design, however, raises several concerns. Most notably, it doubles the number of parameters compared to the baseline feedforward model, making it unsuitable for deployment on extreme-edge devices. Additionally, PCNs do not incorporate the adaptive plasticity of the brain, which efficiently modulates computational effort based on input complexity. As a result, these models may perform unnecessary computations on simpler inputs. 
To overcome these limitations, it is crucial to leverage the recursive nature of predictive coding to construct shallower networks while integrating a dynamic mechanism that adjusts the number of cycles, and consequently, the number of operations, based on input complexity.

To introduce dynamicity into PCN, we leverage early exiting, a well-established technique within the broader framework of Dynamic Neural Networks \cite{laskaridis_adaptive_2021, dynn_survey}. Dynamic networks adjust their computational paths or model parameters at runtime based on the complexity of the input sample \cite{dynamic_routing}. Early exiting, in particular, enables adaptive inference by integrating lightweight decision blocks at intermediate layers of a backbone model. These blocks make per-input halting decisions, allowing the model to return a response if a given confidence threshold on a quality metric (e.g., class probability) is met. Otherwise, computation continues through deeper layers for finer processing \cite{ee_survey}. This mechanism naturally distinguishes between "easy" and "hard" samples—easy samples exit earlier, while harder ones traverse the full network for more thorough analysis. Incorporating early exiting into PCN enables us to determine when to halt further PC cycling over the backbone once a satisfactory response is reached. This not only reduces computational complexity but also aligns with inhibitory mechanisms observed in the brain \cite{luo_architectures_2021}.


Our contributions can be summarized as follows:
\begin{itemize}

\item We propose a new derivation of PC cycling rules in the context of bidirectional PCN that effectively implements feedback and feedforward update rules.

\item We leverage PC dynamics to design shallow PCN that achieve accuracy on par with deeper networks while substantially reducing the {\em memory footprint}, making them well-suited for deployment on extreme-edge devices.

\item We improve the efficiency of PCN inference by introducing a dynamic early-exiting mechanism, allowing for adaptive adjustment of the number of cycles.
\item We utilize knowledge distillation across cycles to enhance the training of early-exiting PCN, thereby improving the performance of early cycles (i.e., exits).


\end{itemize}

\section{Proposed Architecture}
\label{arch}

In this section, we introduce our proposed PCN model enhanced with early exits and outline the inference process and training methodology.

\subsection{Early Exiting PCN Architecture}
\label{early_exit}

\begin{figure}[tb]
    \centering
    \includegraphics[width=0.85\columnwidth]{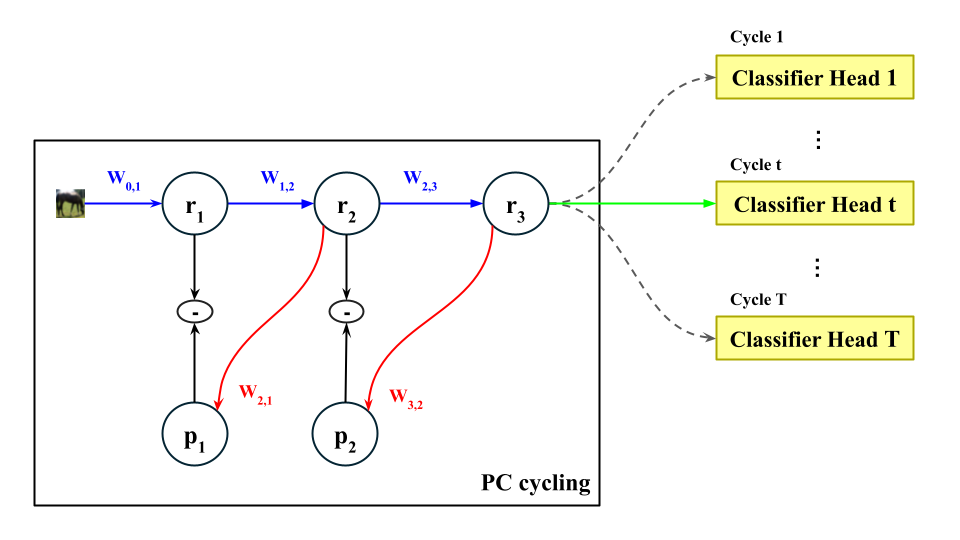}
    \caption{Proposed model --- PCN appended with early exiting classifiers. (Blue: input data to the model, Red: PC cycling of feed-forward and backward layers to reduce local errors, Green: exit classifier for cycle $t$)}
    \label{fig:arch}
    \vspace{-0.5cm}
\end{figure}

Our proposed model is illustrated in \Cref{fig:arch}. It consists of a shared backbone serving as a feature extractor, along with downstream task classifiers. The backbone is designed as a bidirectional hierarchy of convolutional and deconvolutional layers, where blue arrows denote the forward convolutional pass and red arrows indicate the feedback deconvolutions. During inference, the model performs a variable number of cycles, $t \leq T$, over the backbone to iteratively minimize local prediction errors across all layers. Once the cycling process concludes, the final layer feature vector is passed to the classifier corresponding to the current cycle count $t$ (green arrow in \Cref{fig:arch}). The classification confidence is then compared against a predefined user threshold. If the confidence exceeds the threshold, the inference is terminated, and a response is returned. Otherwise, another cycle is initiated, followed by another classification and threshold comparison. 

In the proposed architecture, the number of classifiers corresponds to the maximum number of allowed cycles, denoted as $T$. The decision to employ $T$ distinct classifiers, rather than a single classifier shared across all cycles, is driven by the evolving nature of feature representations throughout the iterative process. Since feature vectors undergo continuous refinement from one cycle to the next, a classifier trained on feature representations from a five-cycle model would be unable to accurately interpret the patterns extracted by a one-cycle model for the same input. In the following subsections, we will further detail the cycling rules as well as the training scheme.


\subsection{PC Cycling Rule}
\label{rules}
We consider a PCN comprising $L$ layers. The forward convolutional weights connecting layer $l$ to layer $l+1$ are denoted by $\mathbf{W}_{l,l+1}$, while the backward (feedback) convolutional weights from layer $l+1$ to layer $l$ are represented as $ \mathbf{W}_{l+1,l}$.  
We define $\mathbf{r}_l(t)$ as the feature representation at convolution layer $l$ and cycle $t$. The representation at layer $l = 0$, i.e. $\mathbf{r}_0(t)$, is fixed for all cycles and is equal to the input image. Further, for $t = 0$, all feature representations are initialized through a standard feedforward pass, given by:
\[
 \mathbf{r}_{l}(0) = \phi(\mathbf{W}_{l-1,l} \mathbf{r}_{l-1}(0)), \qquad l=1,\cdots,L
\]
where \( \phi \) a nonlinear activation function, which we assume to be 
ReLU in our experiments. 

To derive the PC update rules, we apply gradient descent to minimize the {\em local errors} at each pass.

\subsubsection{Feedback pass update}
The feedback pass update rule governs a process in which the higher-layer representation, $\mathbf{r}_{l+1}(t)$, generates a top-down prediction of the lower-layer representation, $\mathbf{r}_l(t)$, denoted by $\mathbf{p}_l(t)$. This prediction is then used to update $\mathbf{r}_l(t)$. The top-down prediction is given by:
\begin{equation*}
    \mathbf{p}_{l}(t) = \phi \left[ \mathbf{W}_{l+1,l} \mathbf{r}_{l+1}(t)\right]
\end{equation*}
The update is  carried out by minimizing the local error, which is defined as:
\begin{equation*}
    \epsilon_{l}(t)=\frac{1}{2} \left|\left|\mathbf{r}_{l}(t)- \mathbf{p}_{l}(t)\right|\right|_2^2
\end{equation*}
where $|.|_2$ denotes the $L_2$-norm. The gradient is found to be:
\begin{equation*}
\frac{\partial \epsilon_l(t)}{\partial \mathbf{r}_{l}(t)}=
\mathbf{r}_{l}(t)-\mathbf{p}_{l}(t)
\end{equation*}
For each layer, we update the representation $\mathbf{r}_{l}(t)$ using gradient descent with a learning rate $\alpha_l$. To maintain notational clarity, and given that each cycle consists of a feedback pass and a feedforward pass, the feedback updates are computed at the midpoints between consecutive cycles. Hence, the feedback update rule, computed as $t+1/2$, with $t=0,\cdots,T-1$, can be expressed, for $l=1,\cdots,L$, as:
\begin{equation}
    \mathbf{r}_{l}(t+1/2)=(1-\alpha_l)\mathbf{r}_{l}(t)+\alpha_l 
    \phi \left[ \mathbf{W}_{l+1,l} \mathbf{r}_{l+1}(t)\right]
\end{equation} 
The representation of the last layer remains unaffected during the feedback pass by design, i.e. $    \mathbf{r}_{L}(t+1/2)=\mathbf{r}_{L}(t)$.

\subsubsection{feed-forward pass update} 
The feed-forward pass update rule governs a process in which the lower-layer representation, generates a bottom-up prediction, which is then used to update the upper-layer representation.
 The feed-forward prediction is given by:
\[
  \mathbf{p}_{l}(t+1/2) =\phi [\mathbf{W}_{l-1,l} \mathbf{r}_{l-1}(t+1/2)]
\]
Similar to the feedback process, we compute the gradient of the prediction error with respect to $\mathbf{r}_{l}(t+1/2)$. This results in the following feed-forward update rule, for $l=1,\cdots,L$:  
\begin{equation}
\mathbf{r}_{l}(t+1)= (1-\beta_l)\mathbf{r}_{l}(t+1/2)+\beta_l \phi [ \mathbf{W}_{l-1,l} \mathbf{r}_{l-1}(t+1/2)]
\end{equation}
where $\beta_l$ is the learning rate for layer $l$.

While drawing inspiration from \cite{wen_deep_2018}, the above derivations differ in key ways. The feedback pass update remains the same, but our feedforward pass update follows a different formulation. Indeed, in \cite{wen_deep_2018}, both feedback and feedforward pass updates rely solely on the feedback convolution weight matrices, $\mathbf{W}_{l,l-1}$, without utilizing the feedforward convolution weight matrices, $\mathbf{W}_{l-1,l}$. This is because, in \cite{wen_deep_2018}, the gradients for both feedback and feedforward passes are derived solely from top-down predictions. In contrast, our formulation integrates both top-down and bottom-up predictions, leading to a more comprehensive update mechanism. 

It is worth pointing out that for any cycle $t$, the last layer feature representation $\mathbf{r}_{L}(t)$ serves as an input for the classifier head.

\subsection{Early Exiting PCN Training}
\label{loss_sec}


As previously mentioned, each cycle in PCN is associated with a classification head. Consequently, the classification task can be formulated as a multi-objective optimization (MOO) problem, where $T$ losses, denoted as $\mathcal{L}_{i}$, $i\in[1,T]$, compete over the shared convolutional and deconvolutional weights of the PCN. Formally, the training problem is expressed as:
\[
    \min \ (\mathcal{L}_{{1}}, \dots, \mathcal{L}_{{T}})
\]
We address the learning problem using scalarization, which transforms the MOO problem into a single-objective optimization problem through an aggregation rule \cite{moo_book}. A widely used approach is linear scalarization \cite{deep_feature_surgery}, where the overall loss is formulated as a weighted average of the individual losses.

Furthermore, we enhance our training strategy by incorporating Kullback-Leibler (KL) divergence, denoted as $\mathcal{KD}$,  between intermediate logits and the final-cycle logits to facilitate knowledge distillation \cite{li_improved_2019, wolczyk_zero_2021}. In this framework, the deepest network (i.e., the last-cycle network) acts as the teacher, while the preceding shallow sub-networks serve as students. This additional regularization helps early-cycle networks learn more complex patterns despite their limited capacity. Consequently, the total loss can be expressed as:
\[
    \mathcal{L}_{\rm tot} = \rho \sum_{i=1}^T \lambda_i \mathcal{L}_{{i}} + (1-\rho)\sum_{i=1}^{T-1} \mathcal{KD}(\mathbf{\hat{y}}_i, \mathbf{\hat{y}}_T)
\]
where $\lambda_i$ is a positive weighting factor for the loss function $\mathcal{L}_{i}$, $\mathbf{\hat{y}}_i$ represents the logit vector from classifier $i$, and $\rho$ is a balancing factor that regulates the contribution of the two terms in the total loss.




\section{Experiments}

In this section, we empirically show that recursive processing with PC update rules in a deployable shallow model on extreme edge devices can deliver competitive performance compared to established models. Furthermore, we illustrate the benefits of incorporating an early exiting mechanism into PCN to minimize redundant computations.

\begin{table}[tb]
\caption{Model configurations}
    \begin{center}
  \begin{tabular}{c|c|c}
        \hline
         A & B & C \\
        \hline
        \multicolumn{3}{c}{Input image 32x32 }\\
        \hline
        Conv-\textbf{16} & Conv-\textbf{16} & Conv-\textbf{32}\\
        \hline
        Conv-\textbf{16} & Conv-\textbf{16} & Conv-\textbf{32}\\
        \hline
        Conv-\textbf{32} & Conv-\textbf{32} & Conv-\textbf{64}\\
        \hline
        Conv-\textbf{32} & Conv-\textbf{32} & Conv-\textbf{64}\\
        \hline
        Conv-\textbf{64} & Conv-\textbf{64} & Conv-\textbf{128}\\
        \hline
        Conv-\textbf{64} & Conv-\textbf{64} & Conv-\textbf{128}\\
        \hline
        -  & Conv-\textbf{64} & - \\
        \hline
        - & Conv-\textbf{64} & - \\
        \hline
        \multicolumn{3}{c}{Global Average Pooling}\\
        \hline
        \multicolumn{3}{c}{$T$ Fully Connected layers}\\
        \hline
\end{tabular}

    \label{tab:models}
    \end{center}
\vspace{-0.5cm}
\end{table}

\subsection{Dataset}

In order to evaluate our method, we choose the CIFAR-10 dataset \cite{krizhevsky_learning_nodate}. It includes 60000 32x32 RGB images evenly distributed over 10 classes, with 6000 images per each. CIFAR-10 is widely adopted by tiny machine learning benchmarks \cite{banbury2021mlperf}, and the images simulate numerous IoT applications with low-resolution cameras (e.g., surveillance for eyewear protection detection and smart farming for fruit disease classification). For model learning, the training set is formed of 50000 images, and the rest is destined for testing. As a data augmentation procedure, we applied random translation and horizontal flipping. The training set was clustered into 128-sized batches.

\subsection{Model configuration}
The model design process was guided by the goal of leveraging PC dynamics to develop shallow networks capable of running on extreme edge devices with memory constraints ranging from kilobytes (KB) to megabytes (MB). The models presented in \Cref{tab:models} are specifically designed to test the robustness of PC feature updates. Starting with a shallow baseline (Model A), we systematically modify its structure: Model B is deepened, and Model C is widened. 
Our proposed models are based on VGG-like architectures, where all convolutions use a 3×3 kernel with a stride of 1 and are followed by a ReLU activation function. Whenever the number of channels changes, we apply either max-pooling (feed-forward direction) or upsampling (feedback direction) with a 2×2 kernel. Finally, the early exit classifiers are simple linear layers to ensure minimal overhead. The additional number of parameters is $T \times C \times 10$, where $C$ is the number of channels in the last convolution. Note that we have $T$ classifiers corresponding to $T$ cycles as $\mathbf{r}_L(t)$ is the input to the classifier $t$.

For comparison, we adopt several baselines: the MLPerf benchmark for tiny image classification models (TinyPerf) \cite{banbury2021mlperf}, SqueezeNext as a representative edge-specific model (Sq.Next) \cite{squeeze}, model C trained without PC cycling rules (Wen-C-FF) and with 6 PC cycles (Wen-C-6) from \cite{wen_deep_2018} to evaluate our proposed PC update rules without early exiting, and VGG-11 \cite{vgg_11} as a baseline for deep networks. We denote our early-exiting predictive coding networks as EE-PCN-$i$, where $i \in \{$A, B, C$\}$, corresponding to the respective model configurations and PCN-C-6, the network with 6 PC cycles following our mathematical derivation without early exiting (i.e., one classifier after cycling 6 times over the backbone).

\subsection{Training \& Evaluation} 

\begin{table}[t]
\caption{Accuracy (\%) of baseline models and average accuracy across cycles for the proposed EE-PCN given a threshold $\tau$}
\centering
\begin{tabular}{c|c|c|c|c}
\hline
Model & $\tau=0.6$ & $\tau=0.7$ & $\tau=0.8$ & $\tau=0.9$ \\
    \hline
    TinyPerf & \multicolumn{4}{c}{85.00 \cite{banbury2021mlperf}} \\
   \hline
   Sq.Next & \multicolumn{4}{c}{86.82 \cite{squeeze}} \\
   \hline
   Wen-C-FF & \multicolumn{4}{c}{88.23 \cite{wen_deep_2018}} \\
   \hline 
   Wen-C-6 & \multicolumn{4}{c}{92.40 \cite{wen_deep_2018}} \\
   \hline
   VGG-11 & \multicolumn{4}{c}{91.30 \cite{vgg_11}} \\
   \hline \hline
   PCN-C-6 & \multicolumn{4}{c}{91.95} \\
   \hline
   EE-PCN-A & 87.34 & 87.42 & 87.41 & 87.44  \\
    \hline
    EE-PCN-B & 88.42 & 88.40 & 88.20 & 88.06  \\
     \hline
   EE-PCN-C & 89.74 & 89.75 & 89.80 & 89.81  \\
    \hline
\end{tabular}
\label{tab:perf}
\vspace{-0.5cm}
\end{table}

For training, we cycle over the PC backbone for a maximum of $T = 6$ cycles. We use mini-batch stochastic gradient descent for optimization, initialized with a learning rate of 0.01 and gradually reduced by a factor of 10 after a patience of 10 epochs. We set the weight decay to $5e^{-4}$ and momentum to 0.9. All models were trained for 300 epochs with early stopping after 50 epochs if no improvement. We assign equal weights to the losses of all individual classifiers, i.e. $\forall i \in [1, T], \lambda_i = 1$, and set $\rho=0.8$. 

For evaluation, we report the average accuracy across cycles, where a given exit accuracy is the product of the number of images that exited at that considered exit and the number of those that are correctly classified. In addition, we assess efficiency using the number of parameters, model size (measured with 32-bit floating-point weights, FP32), and the number of operations (FLOPs).

\section{Discussion}
\label{discussion}

From a model performance perspective, \Cref{tab:perf} presents the accuracy of our baseline models and proposed PCN models. Despite increasing thresholds, the average accuracy of EE-PCN models remains largely unaffected. This can be attributed to the fact that higher thresholds push more images to exit at later stages. These later exits correspond to a higher number of cycles, allowing the model to correctly classify difficult images with higher confidence.
Since our reported accuracy depends on the product of the number of exited images and the proportion of those correctly classified, even a small number of correctly classified images contributes to improving overall accuracy. Our results thus support the idea that additional cycles enhance the expressivity of shallow models, improving their ability to learn complex patterns. This is particularly beneficial when distinguishing between difficult classes, as it enables the model to extract more distinctive features.
Furthermore, our proposed models outperform our edge-specific baseline models while achieving accuracy closer to that of VGG-11, but with significantly fewer parameters, as shown in \Cref{tab:charac}. Finally, we can appreciate the benefit of utilizing PC in conventional CNN, as we perform better than the equivalent feed-forward CNN (Wen-C-FF vs. PCN-C-6) and highlight the fact that our derived PC rules can achieve comparable results with those of \cite{wen_deep_2018} while combining both top-down and bottom-up predictions.


\begin{table}[t]
\caption{Number of parameters and Memory footprint of \\ baseline and proposed EE-PCN models}
    \centering
    \begin{tabular}{c|c|c}
    \hline
        Model & \#Params ($10^6$) & Size (MB) \\
        \hline
        VGG-11 & 28.15 & 107.4    \\
        \hline
        Sq.Next & 0.59 & 2.25    \\
        \hline \hline
        EE-PCN-A & 0.15 & 0.56   \\
        \hline
        EE-PCN-B & 0.30 & 1.13   \\
        \hline
        
        EE-PCN-C & 0.58  & 2.22   \\
        \hline
    \end{tabular}
    \label{tab:charac}
\end{table}

\begin{figure}[t]
\centerline{\includegraphics[width=0.85\columnwidth]{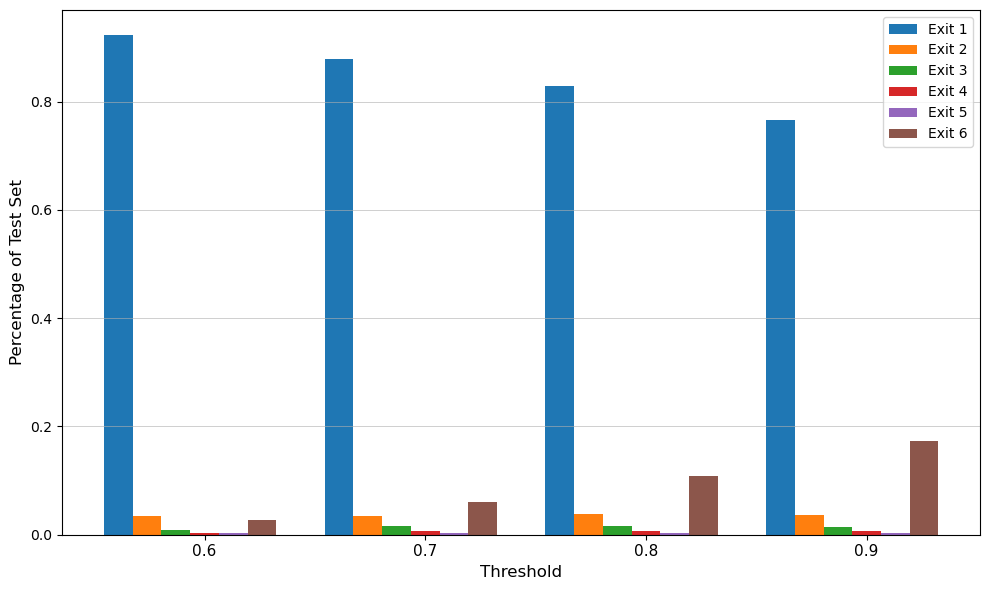}}
\caption{Ratio of test images exiting at different cycles with respect to a given confidence threshold for Model A}
\label{thres}
\vspace{-0.5cm}
\end{figure}

Furthermore, \Cref{thres} highlights the advantage of integrating an early exiting mechanism into PCN. We analyze EE-PCN-A and plot the number of released images for different confidence thresholds. With a confidence threshold of 70\%, EE-PCN-A releases around 85\% of the test data (i.e., 8500 images) while requiring approximately $0.3 \times 10^8$ FLOPs per image. This results in an overall 82.86\% reduction in FLOPs compared to static execution with VGG-11. Thus, when a model exits early (i.e., at $T = 1$ or $T = 2$), it not only benefits from a small memory footprint but also achieves a significantly lower FLOP count than deep networks. This leads to substantial energy savings, enabling battery-powered extreme-edge devices to operate for extended periods.

\begin{figure}[t]
\centerline{\includegraphics[width=0.85\columnwidth]{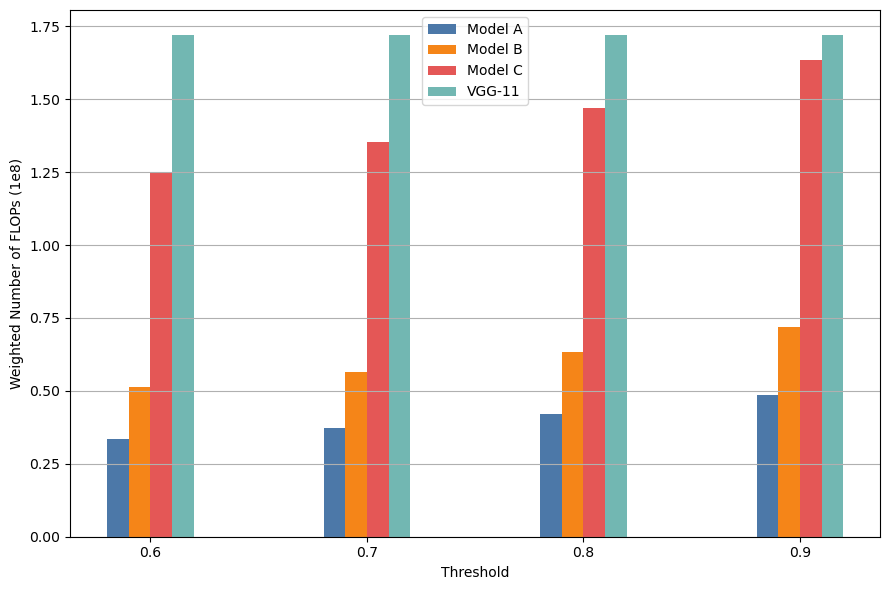}}
\caption{Weighted number of FLOPs for the proposed EE-PCN models against VGG-11}
\label{FLOPs}
\vspace{-0.5cm}
\end{figure}

Since our goal is to deploy our models on extreme edge devices, memory footprint plays a crucial role in our design. In \Cref{tab:charac}, we clearly see that the proposed models meet the memory constraints of many extreme edge devices with only a few megabytes of storage. Notably, EE-PCN-A can be further compressed using 8-bit integer weights, reducing its size to approximately 143 KB, which fits within the memory limits of most frugal variants of STM32 microcontrollers.
However, the recursive nature of PCN introduces additional numerical operations, which can be computationally demanding. To better analyze this trade-off, we plot in \Cref{FLOPs} the required FLOPs for processing all test set images as a function of the number of cycles. We observe that, up to a certain model-dependent number of cycles, the computational cost remains below that of VGG-11. Indeed, numerous input images do not require complex processing, resulting in a significant reduction in compute demand from the edge device. 

\section{Conclusion}

In this paper, we proposed a shallow network for image classification based on PC dynamics and early exiting for extreme edge devices. We found that PC can yield high accuracy without resorting to deep models. Moreover, since PC cycling demands a high number of cycles, we employed early exiting to abort further computation once a user-predefined performance target is reached. Finally, we could achieve competitive results with deep networks while drastically reducing the memory footprint. Our future work will be to further optimize the training process by quantifying the individual impacts of PC cycling, linear scalarization, and knowledge distillation on the overall performance and propose novel methodologies to combine them efficiently. Moreover, we intend to assess our approach to real-world edge applications where large images are often used, such as video surveillance and traffic monitoring.

\bibliographystyle{IEEEtran}
\bibliography{refs}

\end{document}